\documentclass[11pt,a4paper]{article}
\usepackage[hyperref]{naaclhlt2019}
\usepackage{times}
\usepackage{latexsym}
\usepackage{url}
\usepackage{wrapfig}
\usepackage{multirow}
\usepackage[T1]{fontenc}
\usepackage{microtype}
\usepackage{graphicx}
\usepackage{subfigure}
\usepackage{booktabs} 
\usepackage{hyperref}
\usepackage{enumitem,xcolor}
\usepackage[normalem]{ulem}
\usepackage[ruled,linesnumbered]{algorithm2e}
\usepackage{tikz}
\usetikzlibrary{fit,positioning, calc}

\tikzset{
    between/.style args={#1 and #2}{
         at = ($(#1)!0.5!(#2)$)
    }
}

\usepackage[utf8]{inputenc} 
\usepackage[T1]{fontenc}    
\usepackage{hyperref}       
\usepackage{url}            
\usepackage{booktabs}       
\usepackage{amsfonts}       
\usepackage{amsmath}
\usepackage{amsthm}
\usepackage{nicefrac}       
\usepackage{microtype}      
\usepackage{color}
\usepackage{graphicx}
\usepackage{tikz}
\usepackage{bbm}
\usepackage{mathtools}
\usepackage{comment}
\usetikzlibrary{arrows.meta,calc,decorations.markings,math}
\usetikzlibrary{bayesnet}
\theoremstyle{plain}

\theoremstyle{definition}

\aclfinalcopy 

\usepackage{listingsutf8}
\usepackage{listings} 
\title{GASC: Genre-Aware Semantic Change for  Ancient Greek}

\author{Valerio Perrone \\
  Amazon, Berlin\Thanks{  Work done prior to joining Amazon.} \\
  {\small {\tt vperrone@amazon.com}} \\
  \And
  Marco Palma \\
  University of Warwick \\
  {\small {\tt m.palma@warwick.ac.uk}} \\
  \\
  \And
  Simon Hengchen \\
   University of Helsinki \\
  {\small{\tt simon.hengchen@helsinki.fi}} \\
  \AND
  Alessandro Vatri \\
  University of Oxford \\The Alan Turing Institute \\
  {\small{\tt avatri@turing.ac.uk}} \\
  \And
  Jim Q. Smith \\
  University of Warwick \\The Alan Turing Institute \\
  {\small{\tt j.q.smith@warwick.ac.uk}} \\
   \And
  Barbara McGillivray \\
  University of Cambridge\\ The Alan Turing Institute \\
  {\small{\tt bmcgillivray@turing.ac.uk}} \\
}

\date{}

\begin{document}
\maketitle
\begin{abstract}
Word meaning changes over time, depending on linguistic and extra-linguistic factors. Associating a word's  correct  meaning in its  historical context is a central challenge in diachronic research, and is relevant to a range of NLP tasks, including information retrieval and semantic search in historical texts. Bayesian models for semantic change have emerged as a powerful tool to address this challenge, providing explicit and interpretable representations of semantic change phenomena. However, while corpora typically come with rich metadata, existing models are limited by their inability to exploit contextual information (such as text genre) beyond the document time-stamp.  This is particularly critical in the case of ancient languages, where lack of data and long diachronic span make it harder to draw a clear distinction between polysemy (the fact that a word has several senses) and semantic change (the process of acquiring, losing, or changing senses), and current systems perform poorly on these languages. We develop GASC, a dynamic semantic change model that leverages categorical metadata about the texts' genre to boost inference and uncover the evolution of meanings in Ancient Greek corpora. In a new evaluation framework, our model achieves improved predictive performance compared to the state of the art.
\end{abstract}

\section{Introduction}
Change and its precondition, variation, are inherent in languages. Over time, new words enter the lexicon, others become obsolete, and existing words acquire new senses. These changes are grounded in cognitive, social, and contextual factors, and can be realized in different ways. For example, in Old English \emph{thing} meant `a public assembly'\footnote{In the remainder of this paper, we use \emph{emphasis} to refer to a word and `single quotes' for any of its senses.} 
and currently it more generally means `entity'.
Semantic change research has a number of practical applications, beyond historical linguistics research, including new sense detection in computational lexicography and information retrieval for historical texts that  allows to restrict a search to certain word senses (e.g. the old sense of the English adjective \emph{nice} as `silly'). To take an example from recent semantic change in English, the verb \emph{tweet} used to be uniquely associated with birds' sounds and has recently acquired a new sense related to the social media platform Twitter. However, in this as in many other cases, the original sense co-exists with the new one, and specific contexts or genres will select one over the other. This is known as synchronic variation, and can be successfully modelled probabilistically, as advocated by several authors (see e.g. \citet{jenset}).
The close relationship between innovation and variation is well-known in historical linguistics, and critical to ancient languages. Indeed, the unavailability of balanced corpora due to the limited amount of data at our disposal makes it crucial for models to explicitly account for confounding variables like genre, so as to enable them to use all existing data. 

To address these challenges, we introduce GASC (\textbf{G}enre-\textbf{A}ware \textbf{S}emantic \textbf{C}hange), a novel dynamic Bayesian mixture model for semantic change. In this model, the evolution of word senses over time is based not only on distributional information of lexical nature, but also on additional features, specifically genre. This allows GASC to decouple sense probabilities and genre prevalence, which is critical with genre-unbalanced data such as ancient languages corpora. The value of incorporating genre information in the model goes beyond literary corpora and historical language data and can be applied to recent data spanning over a period of time where text type information is critical, for example in specialized domains. Explicitly modelling genres also makes it possible to address a number of additional questions, revealing the genre most likely associated to a given sense, the most unusual sense for a genre, and which genres have the most similar senses. Naturally, this framework can be applied to different categorical metadata about the text, such as author, geography, or style.
 
\begin{figure*}[t]
\centering
\includegraphics
[height=8cm, width=1.3\columnwidth,trim={0.5cm 0cm 0 0},clip]{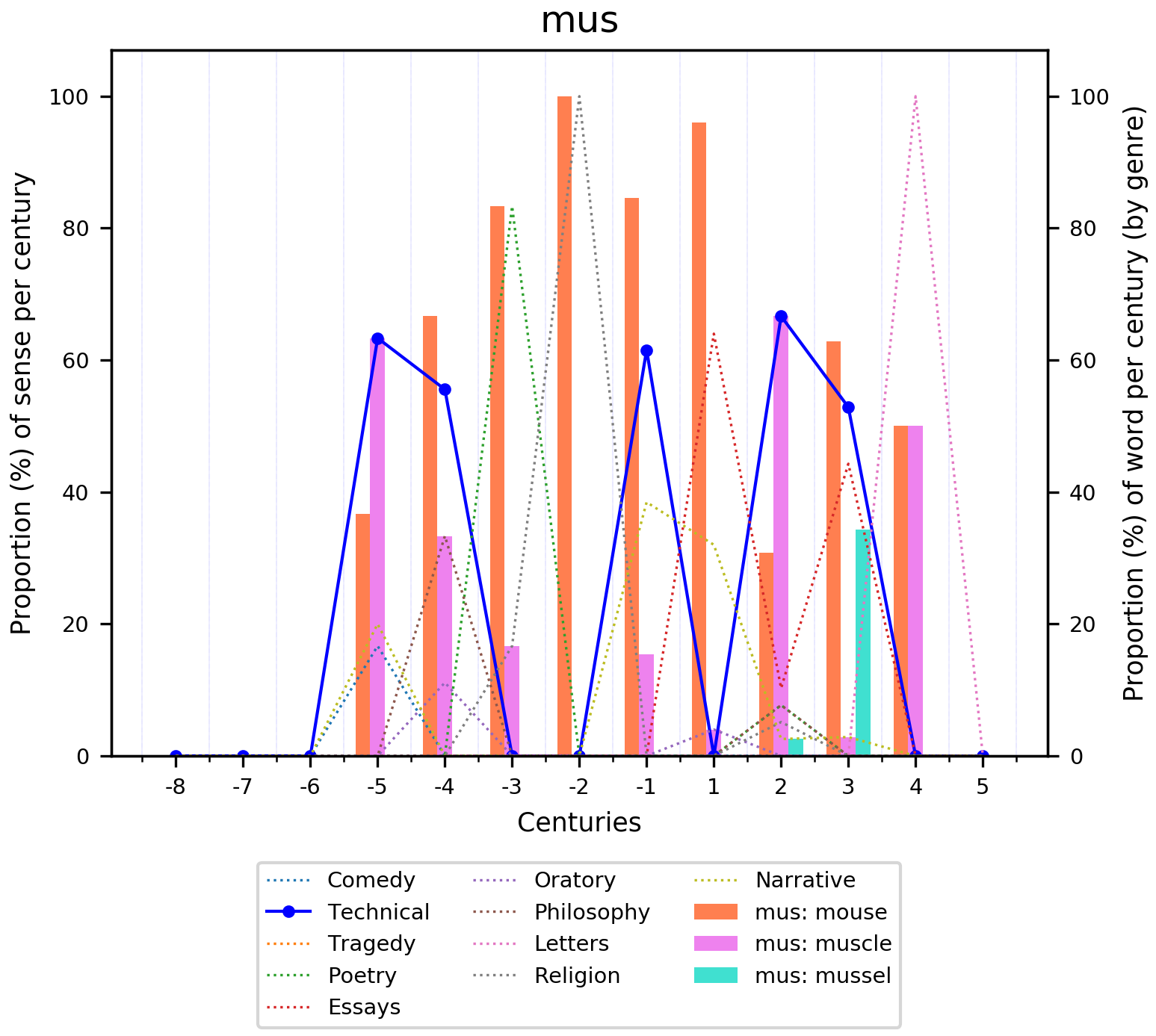}
\caption{Distribution  of \emph{mus} `mouse'/`muscle'/`mussel' by genre vs its senses over time. Lines track \emph{mus} proportions in each genre and century, while bars show the \emph{mus} occurrence proportions with each sense and century.}
\label{musvar}
\end{figure*}

Ancient Greek is an insightful test case for several reasons. First, Ancient Greek words tend to have a particularly high number of senses \citep{bakker_register_2010}, and Ancient Greek texts display a large number of literary genres. Second, we can use data spanning several centuries. Third, Ancient Greek  scholarship  provides high-quality data to validate automatic systems. Top-quality transcribed Ancient Greek texts are available, eliminating the need for OCR correction. 
Finally, polysemous words are particularly sensitive to register variation and the distribution of senses can vary greatly across registers
\citep{Leiwoetal2012}. 
As most extant texts are literary and relatively conservative from a linguistic perspective, we expect genre (the type of a text) and register (the fact that different varieties of language are used in particular situations) to play a significant role in the variation of sense distributions in polysemous words. The word \emph{mus}, for instance, can mean `mouse', `muscle', or `mussel'. The effect of genre on the distribution of its meaning can be estimated visually from Figure \ref{musvar}. In this graph, lines represent the percentage of the occurrences of the target word in a literary genre across centuries, while bars represent the percentage of the occurrences of a specific sense of the target word across centuries. If any line shows a similar trend to that of any set of bars, we may estimate that genre might play a more decisive role than diachrony in determining variation in the distribution of senses. Here, the distribution of 'muscle' over time (pink bars) closely follows the distribution of this word in technical genres over time (blue line), suggesting that the effect of genre should be incorporated into semantic change models.

\section{Related work}
\label{sec:related}
Semantic change in historical languages, especially on a large scale and over a long time period, is an under-explored, but impactful research area. Previous work has mainly been  qualitative in nature, due to the complexity of the phenomenon (cf. e.g. \citet{Leiwoetal2012}. In recent years, NLP research has made great advances in the area of semantic change detection and modelling (for an overview of the NLP literature, see \citet{tang2018state} and \citet{tahmasebi2018survey}), with methods ranging from topic-based models \citep{boydgraber2007, cook2014novel,lau2014learning,wijaya2011understanding, frermann2016bayesian}, to graph-based models \citep{mitra2014s,mitra2015automatic,tahmasebi2017finding}, and word embeddings \citep{kim2014temporal,Basile2018,kulkarni2015statistically,hamilton2016diachronic,dubossarsky2017outta,tahmasebi2018study,rudolph2018dynamic,jatowt2018every,dubossarsky2019timeout}.
However, such models are purely based on words' lexical distribution information and do not account for language variation features such as text type because genre-balanced corpora are typically used. 

With the exception of \citet{Bamman2011} and \citet{Rodda2017}, no previous work has focussed on ancient languages. Recent work on languages other than English is rare but exists: \citet{falk2014quenelle} use topic models to detect changes in French and \citet{hengchen2017does} uses similar methods to tackle Dutch. 
\citet{cavallin2012automatic} and \citet{tahmasebi2018study} focus on Swedish, with the comparison of verb-object pairs and word embeddings, respectively.
\citet{zampieri2016modeling} use SVMs to assign a time period to text snippets in Portuguese, and \citet{tang2016semantic} work on Chinese newspapers using S-shaped models. Most work in this area focusses on simply detecting the occurrence of semantic change, while  \citet{frermann2016bayesian}'s system, SCAN, takes into account synchronic polysemy and models how the different word senses evolve across time.

Our work bears important connections with the topic model literature. The idea of enriching topic models with document-specific author meta-data was explored in \citet{Rosen-Zvi2004} for the static case. Several time-dependent extensions of Bayesian topic models have been developed, with a number of parametric and nonparametric approaches \citep{bleilafferti, gamma, xing, topictime, perrone2017}. In this paper, we transfer such ideas to semantic change, where each datapoint is a bag of words associated to a single sense (rather than a mixture of topics). Excluding cases of intentional ambiguity, which we expect to be rare, we can safely assume that there are generally no ambiguities in a context, and each word instance maps to a single sense. 

\section{The model}\label{sec:model}
We start with a lemmatized corpus pre-processed into a set of text snippets, each containing an instance of the word under study (referred to as ``target word'' in the remainder). Each snippet is a fixed-sized window $W$ of $5$ words to the left and right of the target word. The inferential task is to detect the sense associated to the target word in the given context, and describe the evolution of sense proportions over time.

\begin{figure*}[t]
\centering
\resizebox {1.2\columnwidth} {!} {
    \begin{tikzpicture}
    \tikzstyle{main}=[circle, minimum size = 10mm, thick, draw =black!80, node distance = 16mm]
    \tikzstyle{connect}=[-latex, thick]
    \tikzstyle{box}=[rectangle, draw=black!100, rounded corners]
    
      \node[main] (z) [] {$z^d$};
      \node[main, fill = black!10] (genre) [below=of z] {$g^d$};
      \node[main, fill = black!10] (w) [right=of z] {$w^{d,i}$};

      \node[main, fill = black!10] (w_tneg) [left=of z] {$w^{d,i}$};
      \node[main] (z_tneg) [left=of w_tneg] {$z^d$};
      \node[main, fill = black!10] (genre_tneg) [below=of z_tneg] {$g^d$};
      
      \node[main] (z_tpos) [right=of w] {$z^d$};
      \node[main, fill = black!10] (w_tpos) [right=of z_tpos] {$w^{d,i}$};
      \node[main, fill = black!10] (genre_tpos) [below=of z_tpos] {$g^d$};

      \node[main] (theta) [above=of z] at ($(z)!0.5!(w)$) {$\phi_g^t$ };
      \node[main] (theta_tneg) [above=of z_tneg] at ($(z_tneg)!0.5!(w_tneg)$) {$\phi_g^{t-1}$};
      \node[main] (theta_tpos) [above=of z_tpos] at ($(z_tpos)!0.5!(w_tpos)$) {$\phi_g^{t+1}$};
      \node[draw=none,fill=none] (void1)[left=of theta_tneg]{};
      \node[draw=none,fill=none] (void2)[right=of theta_tpos]{};
      
      \node[main, fill = white!100] (kappa_phi) [above=of theta] {$\kappa^\phi$};
      \node[dashed, circle, draw=black, fill = white!100] (gamma) [left=of kappa_phi] {$a,b$};

      \node[main] (psi_t) [below=1.2in of genre] at ($(genre)!0.5!(w)$) {$\psi_k^t$};
      \node[main] (psi_tneg) [below=1.2in of genre_tneg] at ($(genre_tneg)!0.5!(w_tneg)$) {$\psi_k^{t-1}$};
      \node[main] (psi_tpos) [below=1.2in of genre_tpos] at ($(genre_tpos)!0.5!(w_tpos)$) {$\psi_k^{t+1}$};
      \node[draw=none,fill=none] (void3)[left=of psi_tneg]{};
      \node[draw=none,fill=none] (void4)[right=of psi_tpos]{};

      \node[dashed, circle, draw=black, fill = white!100] (kappa) [below=of psi_t] {$\kappa^\psi$};

      \path (gamma) edge [connect] (kappa_phi)
      		(kappa_phi) edge [connect] (theta)
      		(kappa_phi) edge [connect] (theta_tneg)
      		(kappa_phi) edge [connect] (theta_tpos)
            (void1) edge [dashed] (theta_tneg)
            (theta_tneg) edge [-] (theta)
      		(theta) edge [-] (theta_tpos)
            (theta_tpos) edge [dashed] (void2)
            
            (theta) edge [connect] (z)
            (theta_tneg) edge [connect] (z_tneg)
            (theta_tpos) edge [connect] (z_tpos)
            
    		(z) edge [connect] (w)
            (z_tneg) edge [connect] (w_tneg)
            (z_tpos) edge [connect] (w_tpos)
    		
            (kappa) edge [connect] (psi_t)
            (kappa) edge [connect] (psi_tneg)
            (kappa) edge [connect] (psi_tpos)
            
            (psi_t) edge [connect] (w)
            (psi_tneg) edge [connect] (w_tneg)
            (psi_tpos) edge [connect] (w_tpos)
    
    		(void3) edge [dashed] (psi_tneg)
            (psi_tneg) edge [-] (psi_t)
      		(psi_t) edge [-] (psi_tpos)
            (psi_tpos) edge [dashed] (void4)
            
            (genre) edge [connect] (z)
            (genre_tneg) edge [connect] (z_tneg)
            (genre_tpos) edge [connect] (z_tpos);

      \node[rectangle, inner sep=0mm, fit= (z) (w), xshift=13mm] {};
      
      \node[rectangle, inner sep=3mm,draw=black!100, fit= (w), label={[anchor=south east]south east:$W$}, xshift=0mm] {};
      \node[rectangle, inner sep=3mm,draw=black!100, fit= (w_tneg), label={[anchor=south east]south east:$W$}, xshift=0mm] {};
      \node[rectangle, inner sep=3mm,draw=black!100, fit= (w_tpos), label={[anchor=south east]south east:$W$}, xshift=0mm] {};
      
      \node[rectangle, inner sep=5mm,draw=black!100, fit= (genre) (w), label={[anchor=south east]south east:$D_t$}] {};
      \node[rectangle, inner sep=5mm,draw=black!100, fit= (genre_tneg) (w_tneg), label={[anchor=south east]south east:$D_{t-1}$}] {};
      \node[rectangle, inner sep=5mm,draw=black!100, fit= (genre_tpos) (w_tpos), label={[anchor=south east]south east:$D_{t+1}$}] {};
      \node[rectangle, inner sep=5mm,draw=black!100, fit= (genre_tneg) (w_tneg)] {};
      \node[rectangle, inner sep=5mm,draw=black!100, fit= (genre_tpos) (w_tpos)] {};
      \node[rectangle, inner sep=3mm,draw=black!100, fit= (void1) (theta) (void2), label={[anchor=south east]south east:$G$}] {};
      \node[rectangle, inner sep=3mm,draw=black!100, fit= (void3) (psi_t) (void4), label={[anchor=south east]south east:$K$}] {};
      \node[rectangle, inner sep=3mm, fit= (z) (w),xshift=12.5mm] {};
    
    \end{tikzpicture}
}

\caption{GASC plate diagram with 3 time periods.}
\label{diagram}
\end{figure*}
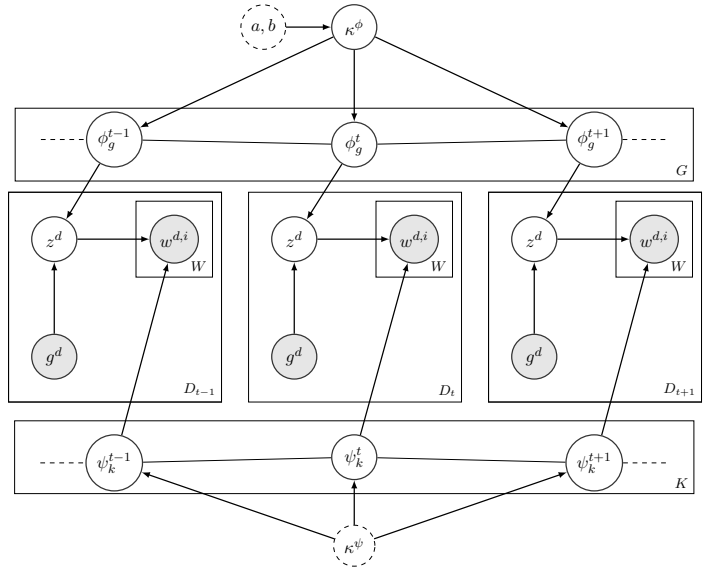

The generative model for GASC is presented in Algorithm~\ref{algo2} and illustrated by the plate diagram in Figure~\ref{diagram}. First, suppose that throughout the corpus the target word is used with $K$ different senses, where we define a sense at time $t$ as a distribution $\psi^t_k$ over words from the dictionary. These distributions are used to generate text snippets by drawing each of their words from the dictionary based on a Multinomial distribution (line 13 in Algorithm~\ref{algo2}). Based on the intuition that each genre is more or less likely to feature a given sense, we assume that each of $G$ possible text genres determines a different distribution over senses (lines 3-4). Each observed document snippet is then associated with a genre-specific distribution over senses $\phi_{g{^d}}^t$ at time $t$, where $g{^d}$ is the observed genre for document $d$. 
Crucially, conditioning on the observed genre we have a specific distribution over senses accounting for genre-specific word usage patterns (line 11). On the other hand, to make sure senses can be uniquely identified across genres, we associate each sense to the same probability distribution over words for all genres. We let word (line 7) and sense distributions (line 4) evolve over time with Gaussian changes, ensuring smooth transitions. 
The coupling between sense probabilities over time is controlled by $K^\phi$, the sense probability precision parameter, so that the larger $K^\phi$, the stronger the coupling between the sense probabilities over time. We place a Gamma prior over $K^\phi$ with hyperparameters $a$ and $b$ (line 1), and infer $K^\phi$ from the data. We fix $K^\psi$, the word probability precision parameter.

\paragraph{Hyperparameter settings}
The model can be applied to different inferential goals: we can focus on the evolution of sense probabilities or on the changes within each sense. For each of these aims, we can use several hyperparameter combinations for $K^\phi$, which is drawn from the prior distribution as determined by $a$ and $b$, and $K^\psi$. Specifically, we consider the following 3 settings. Setting 1: $a = 7$, $b = 3$, $K^\psi = 10$, as in \citet{frermann2016bayesian}. Setting 2: $a = 7$, $b = 3$, $K^\psi = 100$. This aims at enforcing less variation within senses over time. Setting 3: $a = b = 1$, $K^\psi = 100$. This still keeps the bag of words stable for each sense, but also induces less smoothing for sense probabilities over time. Setting 3 allows probabilities to vary widely across centuries. We also expect a large $K^\psi$ to reduce the likelihood of dramatic changes within the same sense across contiguous time periods, and to favour the emergence of new senses. If not otherwise specified, we use setting 3. Other settings (like setting 3 with $K^\psi = 10$) are not recommended since allowing relevant changes over time both in sense probabilities and bag of words might harm interpretability.  A final parameter is the window size $W$, namely the number of words surrounding an instance of the target. While larger windows increase the range of captured dependencies, noise can be introduced in the form of irrelevant contextual words. As in SCAN, we fixed the window size $W$ to 5 for all methods.

\begin{algorithm}
\caption{GASC generative model}
\label{algo2}
\SetAlgoLined

Draw $K^\phi \sim Gamma(a,b)$; \\
\For{\text{time} $t = 1,\dots,T$}{

	\For{genre $g = 1,\dots,G$}{
		Draw sense distribution $\phi^t_g \mid \phi^{-t}_g, K^\phi  \sim N( \frac{1}{2}(\phi^{t-1}_g + \phi^{t+1}_g), K^\phi)$
	}
	
	\For{sense $k = 1,\dots,K$}{
		Draw word distribution $\psi^t_k \mid \psi^{-t}, K^\psi  \sim N( \frac{1}{2}(\psi^{t-1}_k + \psi^{t+1}_k), K^\psi)$ \\
		
	}
	\For{document $d = 1,\dots,D_t$}{
	Let $g^d$ be the observed genre; \\
	Draw sense $z^d \mid g^d \sim \text{Mult(softmax}(\phi^t_{g^d}))$; \\
	
		\For{context position $i = 1,\dots, W$}{
		Draw word $w^{d,i} \sim \text{Mult(softmax}(\psi^{t,z^d}))$; \\
		}
	}
}

\end{algorithm}

\paragraph{Posterior inference}
For posterior inference, we extend the blocked Gibbs sampler proposed in \citet{frermann2016bayesian}. The full conditional is available for the snippet-sense assignment, while  to  sample  the  sense and word distributions we adopt the auxiliary variable approach from \citet{Mimno2008}. The sense precision parameters are drawn from their conjugate Gamma priors. For the distribution over genres we proceed as follows. First, sample the distribution over senses $\phi^t_g$ for each genre $g=1,\dots,G$ following \citet{Mimno2008}. Then, sample the sense assignment conditioned on the observed genre from its full conditional:
$p(z^d \mid g^d, \textbf{w}, t, \phi, \psi) \propto p(z^d \mid g^d, t) p( \textbf{w} \mid t, z^d) = \phi^t_{g} \prod_{w \in \textbf{w}} \psi^{t,z^d}_w.$
This  setting easily extends to sample genre assignments for tasks where, for example, some genre metadata are missing.

\section{Evaluation framework} \label{sec:evaluation}
Evaluating models tackling lexical semantic change is notoriously challenging. Frameworks are either lacking or focus on very specific types of sense change \citep{schlechtweg2018durel,tahmasebi2018survey}. Exceptions are \citet{kulkarni2015statistically}, \citet{Basile2018} and \citet{hamilton2016diachronic}, who focus on the change points of word senses. However, in the case of Ancient Greek (and other historical languages), corpora typically contain gaps and uneven distribution of text genres, and semantic change is so closely related to polysemy that it is hard to find a specific point in time when a new sense emerged in the language. Therefore, it is more appropriate to take a probabilistic approach to model sense distribution, and devise an evaluation approach that fits this.
Although historical dictionaries and traditional philology describe the evolution of word senses over time, they do not necessarily reflect the evidence from corpora on which models can be evaluated, and often only provide insights into the appearance of a new sense, rather than the relative predominance of a word's senses across time. These reasons led us to craft a novel evaluation dataset and framework, which reflects the data on which the model is evaluated, and allows for a finer-grained evaluation of the predominance of word senses across time.

\subsection{Ancient Greek corpus}
\label{sec:data}
We used the Diorisis Annotated Ancient Greek Corpus 
\citep{diorisis2018}, consisting of 10,206,421 
lemmatized and part-of-speech-tagged words. The corpus contains 820 texts spanning between the beginnings of the Ancient Greek literary tradition (8\textsuperscript{th} century BC) and the 5\textsuperscript{th} century AD. The corpus covers a number of Ancient Greek literary and technical genres: poetry (narrative, choral, epigrams, didactic), drama (tragedy, comedy), oratory, philosophy, essays, narrative (historiography, biography, mythography, novels), geography, religious texts (hymns, Jewish and Christian Scriptures, theology, homilies), technical literature (medicine, mathematics, natural science, tactics, astronomy, horsemanship, hunting, politics, art history, rhetoric, literary criticism, grammar), and letters. 
In technical texts, we expect polysemous words to have a technical sense. On the other hand, in works more closely representing general language (comedy, oratory, historiography) we expect words to appear in their more concrete and less metaphorical senses; we cannot assume that this distribution holds in a number of other genres such as philosophy and tragedy. 
Whilst genre-annotated corpora are not especially common in NLP, where most tasks rely on specific genres (e.g. Twitter) or on genre-balanced corpora such as COHA \citep{davies2002corpus}, they are more prevailing within humanities, and especially classics. 
Additionally, research on automated genre identification has been flourishing for decades (e.g. \citet{kessler1997automatic}), making the need for genre information in a potential corpus not as much of a hindrance as can be thought. 

\subsection{Log-likelihood evaluation}
\label{subsec:loglik}
First, we compared GASC with the state-of-the-art (SCAN) in terms of held-out data log-likelihood. We chose 50 targets that could be identified as polysemous (e.g. the verb \emph{leg\={o}}, whose senses are `gather' and `tell') based on two criteria: high frequency and a a suitably clear-cut range of meanings. We initially based our selection on the secondary literature and chose 17 words from the well-studied vocabulary of Ancient Greek aesthetics \citep{pollitt_ancient_1974}. We complemented this selection with the inclusion of the 33 most frequent clearly polysemous words identified by an Ancient Greek expert in a frequency-ranked word list extracted from the Diorisis corpus.
The necessity to identify manually suitable words led us to limit their number to 50. For each one of these target words, we randomly divided the corpus into a train (80\%) and test set (20\%). Results on the 50-word dataset are in Section \ref{sec:results}.

\subsection{Expert annotation}
\label{subsec:senselabel1}

To evaluate our method against ground truth, we proceeded as follows. We selected three three target words (\emph{mus} `mouse'/`muscle'/`mussel' and \emph{harmonia} `fastening'/`agreement'/`musical scale, melody', and \emph{kosmos} `order'/`world'/`decoration') based on their frequency and clear-cut polysemy, as indicated by the standard scholarly Ancient Greek-English dictionary \citep{Liddell:1996} and traditional philological scholarship on their semantics (\citealt{pollitt_ancient_1974} on \emph{harmonia} and \emph{kosmos}). These words are especially suitable for an exploratory case study because they exhibit an abstract sense and a concrete counterpart in general, non-technical vocabulary, and are attested in most of the time periods covered by the corpus and across different literary genres. 
Two Ancient Greek experts manually annotated the whole corpus by tagging the senses of the target words in context. One expert selected the correct sense for each occurrence of \emph{mus} and \emph{harmonia}, and the other expert performed the same task on \emph{kosmos}. The results of each expert's annotation task were not reviewed by the other expert (\citealt{annodata} for the dataset).
Table \ref{table:annot} shows an example from the annotated dataset for the word \emph{kosmos}.
The annotators also marked when the semantic annotation was purely based on the target word context, which is the evidence on which the model can rely (category ``collocates''). Only annotations based on collocates were retained in the evaluation. Using this information, the relative frequency of each sense for each target word in any time slice becomes computable, and was used to create ground-truth data on the diachronic predominance of a word's senses as reflected in the corpus.

\begin{table*}
\centering
\scriptsize
\begin{tabular}{|l|l|l|l|l|l|}
\hline
date&	genre&	author&	work&	target word&	sense id\\\hline
-335&	Technical&	Aristotle&	De Mundo&	kosmos&	kosmos:world\\\hline
\end{tabular}
\caption{Example from annotated dataset displaying \emph{Tou de sumpantos ouranou te kai kosmou sphairoeidous ontos kai kinoumenou kathaper eipon}
(``The whole of the heaven, the whole cosmos, is spherical, and moves continuously, as I have said''),
containing the target \emph{kosmos} and its expert-assigned sense `world' (date: 335 B. C.).
}
\label{table:annot}
\end{table*}
\normalsize

\subsection{Automatic sense labelling}
\label{subsec:senselabel2}
For every time period $T$, inferred sense $k$, and genre $G$, GASC outputs a distribution of words with associated probabilities. For instance, the output for \emph{kosmos} (`order', `world', or `decoration') in oratory at time 0 includes:

\begin{scriptsize}
\begin{lstlisting}
T=0,k=0: 
aêr (0.069); mousikos (0.059); gê (0.056); harmonia (0.034); ouranos (0.033); logos (0.030); gignomai (0.021); sphaira (0.021); pselion (0.020); apaiteô (0.019);
T=0,k=1: 
polis (0.035); asebeia (0.014); politeia (0.012); proteros (0.012); naus (0.012); pentêkonta(0.011); aei (0.011); hama (0.011) ; peripeteia (0.011); oikia (0.011).
\end{lstlisting}
\end{scriptsize}

These distributions can be interpreted by experts based on the meanings of the words they group and thus associated to the senses of the target word. Here, $K=0$ includes \emph{aêr} (`air'), \emph{gê} (`earth'), \emph{ouranos} (`sky'), and \emph{sphaira} (`sphere, globe'), which point to the meaning of \emph{kosmos} as `world'. The list for $K=1$ includes \emph{polis} (`city'), \emph{asebeia} (`impiety'), \emph{politeia} (`constitution'), and \emph{oikia} (`household'), which point to the meaning of \emph{kosmos} as `order'. 
On the other hand, the expert annotation provides lists of corpus occurrences of the target word, each associated to a sense label. In Table \ref{table:annot}, the sense label is `kosmos-world' and we can associate lemmas such as \emph{ouranos} `sky' and \emph{sphairoeides} `spherical' to this sense, as these lemmas occur in the corpus context of this target word.

To evaluate against expert annotation, we automatically match the word senses assigned by the annotators (denoted by $s$) with the senses outputted by the model (denoted by $k$). To achieve this, we first measured how closely each model sense $k$ matches each expert sense $s$. We assigned a confidence score to every possible $(k,s)$ pair by comparing the words associated to $k$ in the model output and the words co-occurring with the target word in the annotated corpus sentences labelled with the expert-assigned sense $s$. For \emph{kosmos} with $k=0$, we compare words from the model output, such as \emph{ouranos} `sky', \emph{gê} `earth', and \emph{sphaira} `sphere' with words from the context of the annotated sentences, such as \emph{sphairoeides} `spherical' and \emph{ouranos} `sky'. 
We then  considered two elements. For words from the model output, we consider the normalized probability with which these words $w_{i}$ are associated to the model sense $k$, i.e. $P(w_{i}|k)$. For \emph{kosmos}, \emph{aêr} `air' is associated to probability 0.069, \emph{gê} `earth' to 0.056, and \emph{ouranos} `sky' to 0.033. For the context words from the annotated data, we consider the degree to which these words are associated to an expert sense. In the example of \emph{kosmos} from Table \ref{table:annot}, this is calculated based on how many different senses a context word like \emph{ouranos} `sky' or \emph{sphairoeides} `spherical' is associated to. To measure this degree of association we define the expert score $m(w_i,s)$ of word $w_i$ as 1 divided by the number of senses assigned by the experts to this word. If the word is associated to only one sense $s$ in the annotated data, its expert score $m(w_i,s)$ will be highest (1); if it is associated to two senses, its expert score is 0.5; if it is not assigned to the sense $s$ by the experts, its expert score $m(w_i,s)$ is 0. Formally, we define the confidence score of a pair of model sense $k$ and expert-assigned sense $s$ as $
\text{conf}(k,s) = \sum_{i} P(w_i|k) * m(w_i,s).
$
The score is highest when $P(w_{i})$ and $m(w_{i},s)$ are highest for all words. In extreme cases, $P(w_{i})$ will be 1 if the model estimated $w_{i}$ to be associated to sense $k$ with probability 1 and $m(w_{i},s)$ is 1 (i.e. $w_{i}$ is only found in contexts labelled as $s$ by the experts). This points to $k$ and $s$ being associated to the same words, and thus being the same sense. The confidence is lowest when $k$ and $s$ do not share words, in which case either $P(w_{i})=0$ or $m(w_{i},s)=0$. In contrast with clustering overlap techniques like purity or rand index, we ensure words with a higher inferred probability and uniquely associated to a sense weigh more. The confidence scores were used to find the best matching pair ($k$, $s$): for every expert sense $s$ we selected the sense(s) $k$ for which $\text{conf}(k,s)$  was higher than the random baseline (1 over the number of expert senses) and higher than the sum of the 2d and 3rd best confidence scores, when possible, or assigned NA when both conditions were not matched. We consider NA as an additional expert sense whenever the expert assigned a sense based on other factors than lexical context.

After matching inferred and expert-assigned senses, we computed precision and recall. For every target word and matched pair $(s,k)$, a word is considered correctly assigned to sense $k$ if it also appeared within a 5-word window of the target word in the expert annotation for $s$. In the example above for \emph{kosmos}, $k=0$ and $s$=`kosmos-world', one such word is \emph{ouranos} `sky' as it appears in the model output for $k=0$ and in the context window of a sentence labelled as `kosmos-world' by the annotators. Moreover, we weighted each word by the inferred probability to account for the different degrees of association of words to senses. Specifically, we defined precision as the ratio between the number of words correctly assigned to $k$, weighted by their respective normalised model-estimated probabilities, and the number of words assigned to $k$ by the model. This metric is based on the distributional hypothesis whereby words occurring in similar contexts tend to exhibit similar meanings. We computed precision after stop word removal, limiting the noise from uninformative contextual words. We defined precision in terms of the words assigned to a sense also appearing within a 5-word window of the target in the expert annotation. Our model, as SCAN, only considers those context words to determine word senses, and for the ground truth evaluation we only retained the cases in which the annotators could disambiguate words based purely on their context. We defined recall as the ratio between the number of all words correctly assigned to $k$ (weighted by their probabilities) and number of words assigned to sense $s$ by the experts (weighted by their expert scores).
For each model, precision and recall scores for each $(s,k)$ pair were averaged and used as final scores.
Since recall directly depends on the number of expert words, the metric can only be used to compare models for a specific target word. While the proposed assessment focusses on dynamic mixture models, it can be generalised to any probabilistic model by considering the posterior probability of the gold word sense.

\section{Experiments}
\label{sec:results}

\paragraph{Predictions on held-out words}
Considering the 50-word dataset described in Section~\ref{sec:evaluation}, we evaluated the predictive performance in terms of log-likelihood of held-out data for SCAN (not using any genre information), GASC-all (GASC with all the $G = 10$ available genres) and GASC-narr (GASC with 2 genres, Narrative vs. non Narrative). Narrative and Technical are such that all 50 words occurred at least once in the training and test sets, and analogous results are obtained when GASC with Technical vs. non Technical. For each model, we compared the 3 hyperparameter settings previously reported, with higher scores indicating that a model is better at explaining unseen data.

\begin{figure}[t]
\centering
\includegraphics[width=\columnwidth]{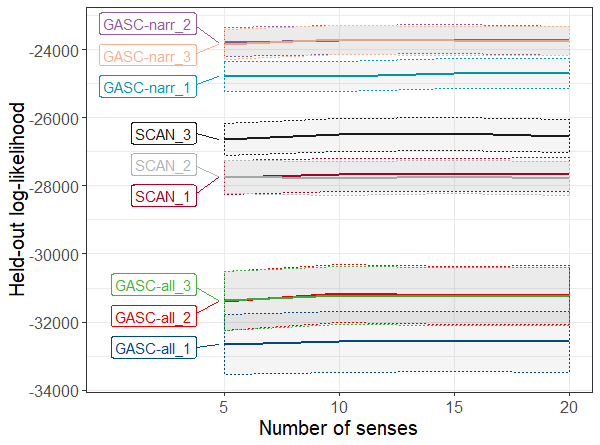}
\caption{Held-out mean log-likelihood varying $K$ (the larger, the better). Shaded areas are $\pm$ 1 standard error. }
\label{loglike}
\end{figure}

Figure~\ref{loglike} shows predictive log-likelihood scores for a range of $K$, with the results averaged over 50 leave-one-out folds. Each time, the scores were averaged under the final 10 samples of the latent variables, out of 1000 MCMC iterations. On average, GASC-narr consistently outperforms SCAN across every $K$ and hyperparameter setting. 
On the other hand, SCAN has a higher held-out log-likelihood than GASC-all. Exploiting some information on the genre yields better predictions, while using all genres attested in the corpus is not effective as some genres are not sufficiently represented by the data.
Figure~\ref{loglike} also shows that the best predictions over unseen data are obtained for $K$ between 10 and 15. Higher $K$ values tend to introduce noisy senses with no improvement for the model output. In addition, Setting 3 worked better or on par with other settings.
In the next section, we fix the hyperparameters and use a validation set of words that were not part of the 50 targets of this experiment.

\paragraph{Ground truth recovery}

\begin{table}
\scriptsize
\addtolength{\tabcolsep}{-2.4pt}
\begin{tabular}{|l|l|l|}
\hline
\multirow{2}{3.5em}{harmonia}&`agreement, harmony'&Technical ($\rho$ = 0.888, p \textless{} 0.0001) \\
&&Narrative ($\rho$ = 0.719, p = 0.006) \\
&&Essays ($\rho$ = 0.561, p = 0.046) \\

\cline{2-3}
&`fastening'&Narrative ($\rho$ = 0.663, p = 0.013)\\

\cline{2-3}
&`stringing,  music scale'&Technical ($\rho$ = 0.817, p = 0.001)\\
&&Philosophy ($\rho$ = 0.632, p = 0.02)           \\
&&Essays ($\rho$ = 0.598, p = 0.031)                \\ 

\hline
\multirow{2}{3.5em}{kosmos}&`decoration'&Narrative ($\rho$ = 0.887, p = 0.001)             \\
&&Technical ($\rho$ = 0.705, p = 0.023)             \\
&&Oratory ($\rho$ = 0.664, p = 0.036)               \\
\cline{2-3}
&`order'&Technical ($\rho$ = 0.875, p = 0.001)             \\
&&Narrative ($\rho$ = 0.862, p = 0.001              \\ 
\cline{2-3}
&`world’&Technical ($\rho$ = 0.792, p = 0.006)             \\
&&Oratory ($\rho$ = 0.723, p = 0.018)               \\ 
\hline
\multirow{2}{3.5em}{mus}&`mouse'&Narrative ($\rho$ = 0.813, p = 0.001)             \\
&& Essays ($\rho$ = 0.743, p = 0.004)                \\ 
\cline{2-3}
&`muscle’&Technical ($\rho$ = 0.766, p = 0.002)             \\ 
\cline{2-3}
&`mussel’& Narrative ($\rho$ = 0.736, p = 0.004)             \\
&& Essays ($\rho$ = 0.736, p = 0.004)                \\
&& Poetry ($\rho$ = 0.613, p = 0.026)                \\
\hline
\end{tabular}
\caption{Correlations between senses and genres for manually annotated target words.}
\label{table:correlations}
\end{table}

\begin{figure*}[ht]
\begin{tikzpicture}

    \draw (-0.5, 5.5) node[inner sep=0] {\includegraphics[trim={0cm 0cm 0 0},clip,scale = 0.15]{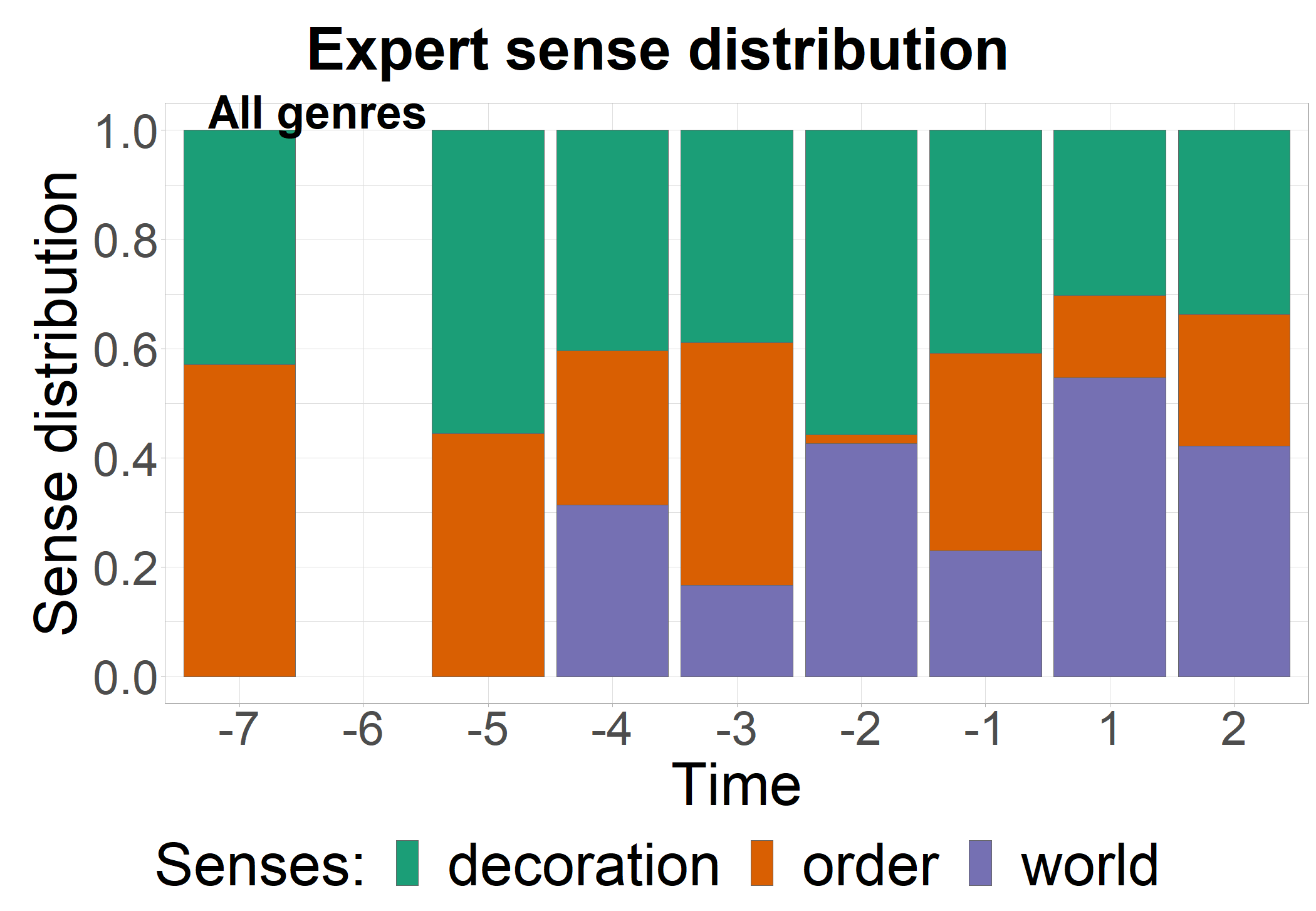}};
    \draw (-3.3, 4.5+2) node {(A)};
    
    \draw (-0.5, 2) node[inner sep=0] {\includegraphics[trim={0cm 8cm 0 0},clip,scale = 0.15, , ]{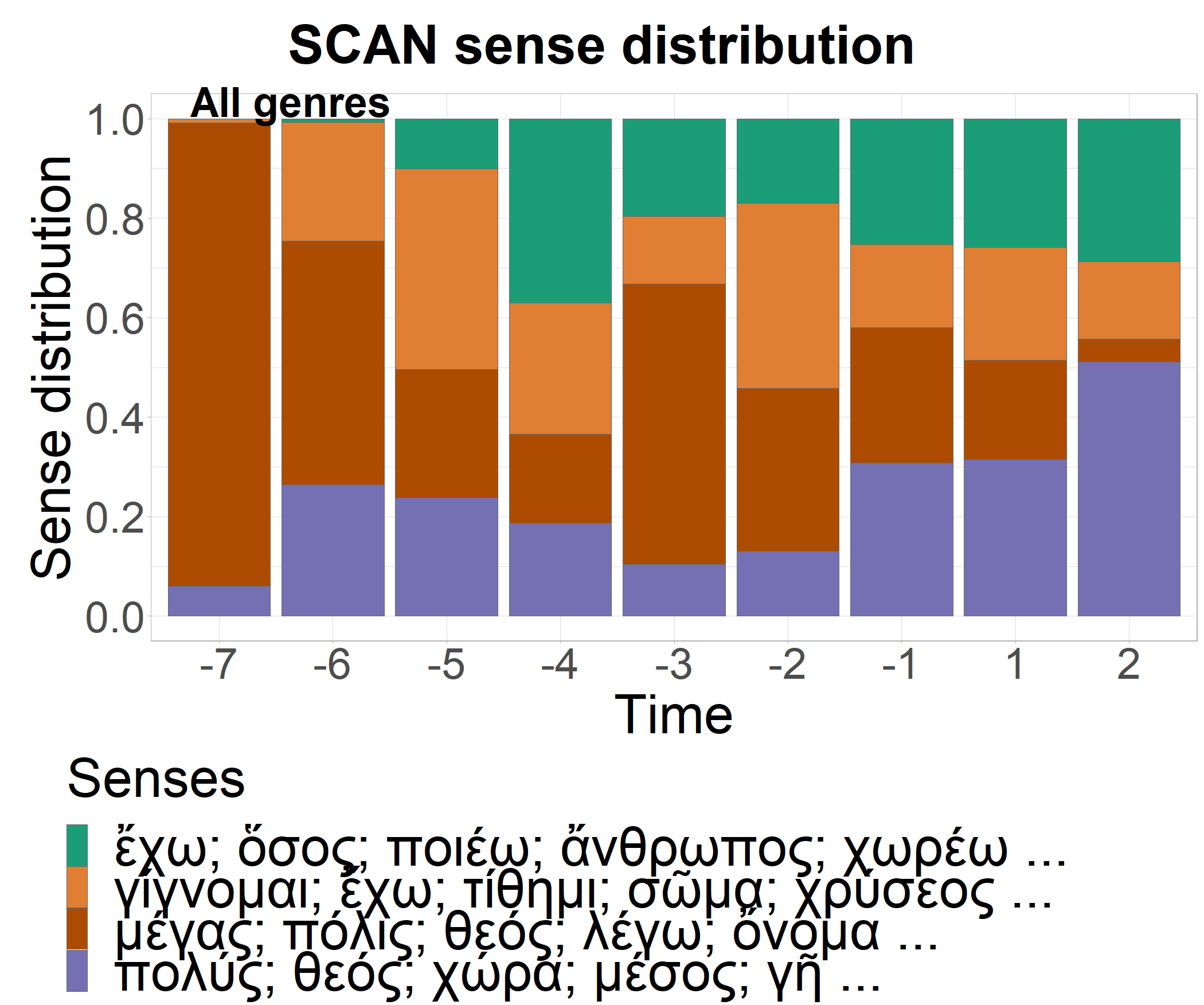}};
    \draw (-3.3, 3) node {(B)};
    
    \draw (7.5, 5.5) node[inner sep=0] {\includegraphics[trim={0cm 0cm 0 0},clip,scale = 0.15]{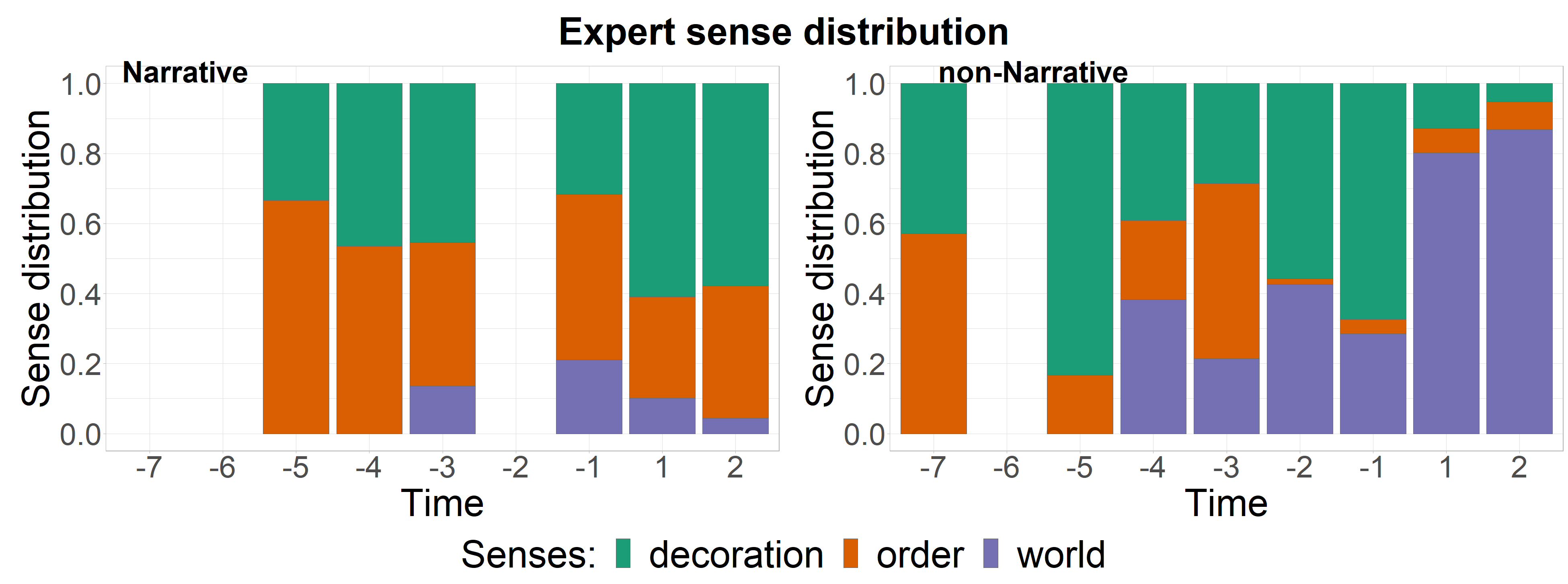}};
    \draw (2.4, 4.5+2) node {(C)};
    
    \draw (7.5, 2) node[inner sep=0] {\includegraphics[trim={0cm 8cm 0 0},clip,scale = 0.15]{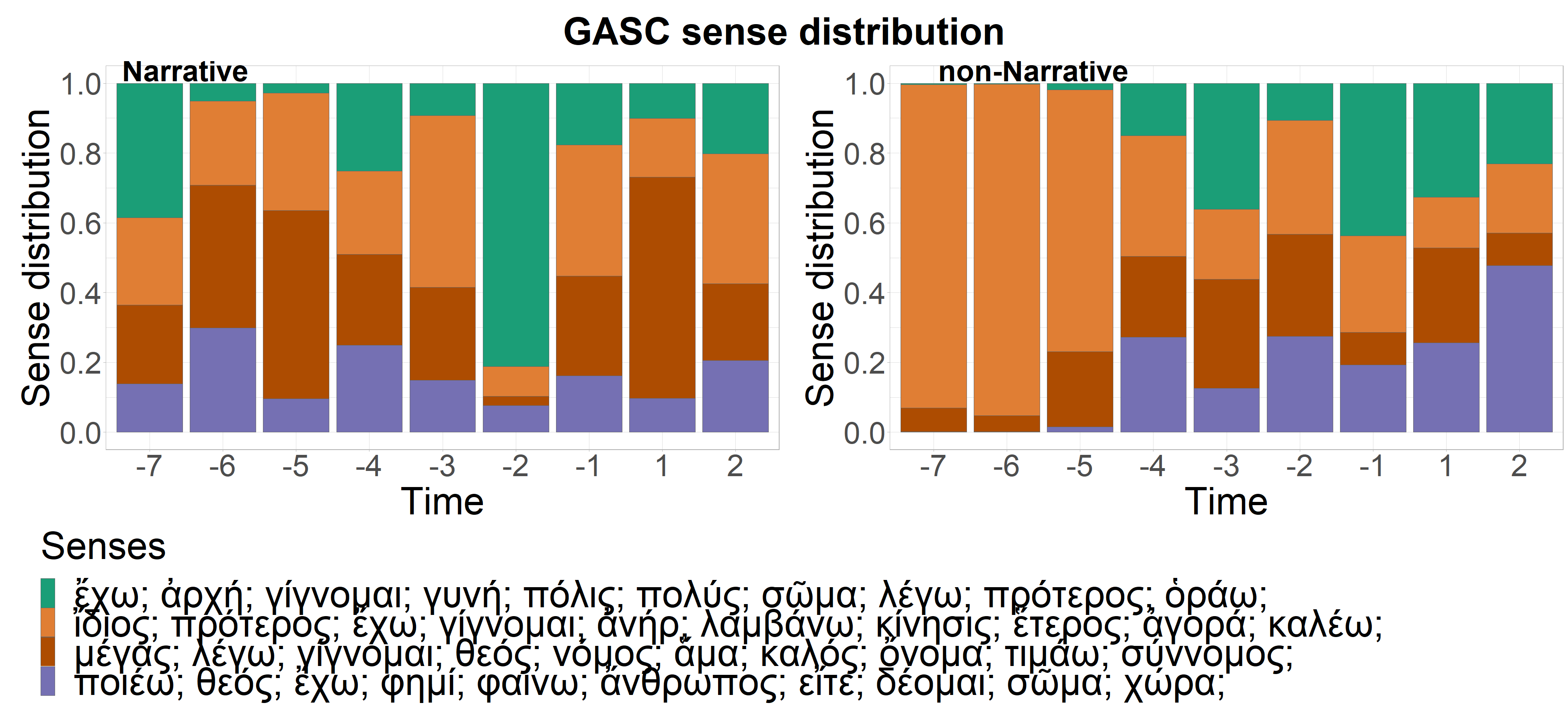}};
    \draw (2.4, 3) node {(D)};
    
\end{tikzpicture}

\caption{Expert annotation (top) vs SCAN and GASC (bottom). Each stacked bar represents all \emph{kosmos} occurrences in a given time. Colours denoting senses are matched between plots. Both shades of orange map to `order', but the fourth sense in (B) and (D) is NA (i.e., $\text{conf}(k,s)$ not higher than the random baseline and not higher than the sum of 2nd and 3rd best confidence scores).
}
\label{barplots}
\end{figure*}
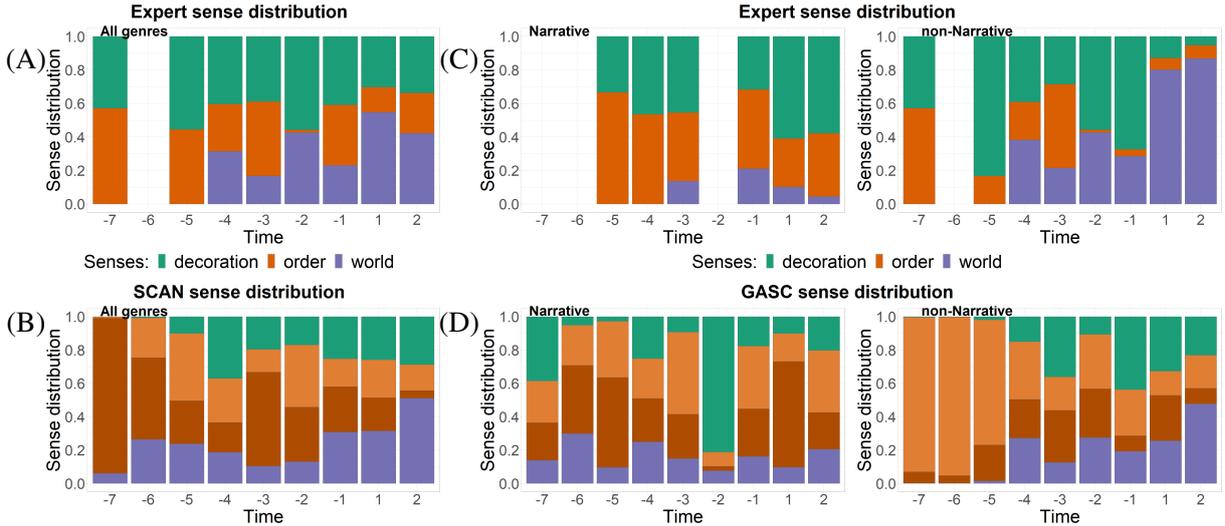

\begin{table*}
\centering
\small

\begin{tabular}{|c||ccc|ccc|ccc|}
\hline

\multicolumn{1}{|c||}{Word/Model}&  \multicolumn{3}{c||}{SCAN} &    \multicolumn{3}{c||}{GASC-independent}  &   \multicolumn{3}{c|}{GASC} \\ \hline
\multicolumn{1}{|l||}{} & \multicolumn{1}{c|}{P} & \multicolumn{1}{c|}{R} & \multicolumn{1}{c||}{F1}
& \multicolumn{1}{c|}{P} & \multicolumn{1}{c|}{R} & \multicolumn{1}{c||}{F1}
& \multicolumn{1}{c|}{P} & \multicolumn{1}{c|}{R} & \multicolumn{1}{c|}{F1}  \\ \hline\hline

\emph{mus}                    & \multicolumn{1}{c|}{\textbf{0.430}}  & \multicolumn{1}{c|}{\textbf{0.477}}  & \multicolumn{1}{c||}{\textbf{0.452}}   & \multicolumn{1}{c|}{0.420}  & \multicolumn{1}{c|}{0.442}  & \multicolumn{1}{c||}{0.431}    & 
\multicolumn{1}{c|}{0.224}  & \multicolumn{1}{c|}{0.298}  & \multicolumn{1}{c|}{0.253}   \\ \hline

\emph{harmonia}               & \multicolumn{1}{c|}{0.527}  & \multicolumn{1}{c|}{0.708}  & \multicolumn{1}{c||}{0.603}   &  \multicolumn{1}{c|}{\textbf{0.582}}  & \multicolumn{1}{c|}{\textbf{0.729}}  & \multicolumn{1}{c||}{\textbf{0.646}}   &  
\multicolumn{1}{c|}{0.497}  & \multicolumn{1}{c|}{0.481}  & \multicolumn{1}{c|}{0.484}   \\ \hline

\emph{kosmos}                 & \multicolumn{1}{c|}{0.405}  & \multicolumn{1}{c|}{0.586}  & \multicolumn{1}{c||}{0.478}   &  \multicolumn{1}{c|}{0.362}  & \multicolumn{1}{c|}{0.447}  & \multicolumn{1}{c||}{0.399}    & 
\multicolumn{1}{c|}{\textbf{0.525}}  & \multicolumn{1}{c|}{\textbf{0.611}}  & \multicolumn{1}{c|}{\textbf{0.595}}  \\ \hline

\end{tabular}
\caption{SCAN vs GASC on \emph{mus} (`mouse', `muscle', `mussel'), \emph{harmonia} (`abstract', `concrete', `musical'), and \emph{kosmos} (`order', `decoration', `world') in terms of precision (`P'), recall (`R'), and F1-score (`F1'). 
}
\label{table:models-overview}
\end{table*}

\normalsize

We explored the ability to recover ground truth when available.
For \emph{mus}, experts annotated 205 instances, of which 198 were assigned to one of the 3 senses `mouse', `mussel', and `muscle'; out of these 198 assignments, 114 were based on lexical contextual information only (category `collocates') and were retained for the evaluation.
For \emph{harmonia}, the number of annotated occurrences was 599, of which 411 were of the type `collocates'. For \emph{kosmos}, 1,411 occurrences were annotated, of which 1,406 were assigned to a sense, and in 1,102 cases the annotation was of the type `collocates'. Only the annotations of the type `collocates' were kept for the expert sense distribution, and thus for the evaluation.
We identified genres with the largest effect on the distribution of senses by the Spearman's Rank Correlation Coefficient for each word-sense $s$ between the frequency $f(s)$ of $s$ across centuries and the frequency $f(s,g)$ of $s$ in each genre $g$ across centuries (Table~\ref{table:correlations}). Significant correlation between $f(s)$ and any $f(s,g)$ suggests that variation in the frequency of a word sense across centuries is not due to diachronic change, but to how frequently $s$ is attested in $g$ in each century (and, ultimately, to the amount of texts representing $g$ in each century). 
Given the amount of available data and the size of the correlations, we considered the genres Technical and non-Technical for \emph{mus} and \emph{harmonia}, and both Technical and non-Technical and Narrative and non-Narrative for \emph{kosmos}. 
These words were selected as examples of polysemous words (a) with a range of clearly distinct senses (such as ‘mus’, whose three senses are strikingly diverse), (b) attested in most, if not all, the time periods covered by the corpus, and (c) attested across a number of genres. As expert annotations of semantic change in Ancient Greek corpora are virtually unavailable, this choice also allowed us to leverage ground truth for validation.

We compared SCAN with GASC and GASC-independent, a simpler version that fits independent models to sets of documents sharing the same genre, so that parameters and senses are inferred independently across genres (while in GASC senses are shared but their probability distributions are independent across genres). First, we compared word senses across time with expert-annotated data. Figure \ref{barplots} shows the time distribution of the senses of \emph{kosmos} in the expert annotation (top) and as outputted by SCAN and GASC run on Narrative vs. non-Narrative (bottom). For non-narrative texts, the GASC sense distribution successfully captures the `world' sense arising only after 400 BC, which is less clear for SCAN. Second, we computed precision, recall, and F1 scores (the harmonic mean of precision and recall) to determine how closely the words assigned to a sense match the ones assigned by experts (Table~\ref{table:models-overview}). For GASC, the values average precision, recall, and F1-score for \{Technical, non-Technical\} for \emph{mus} and \emph{harmonia} and \{Narrative, Non-Narrative\} for \emph{kosmos}. 
The results show that, for the most represented targets, genre information improves the ability to recover the ground truth.  

\section{Conclusion}\label{sec:discussion}
We introduced GASC, a Bayesian model to study the evolution of word senses in ancient texts. We performed this analysis conditional on the text genre, demonstrating that the ability to harness genre metadata addresses a fundamental challenge in disambiguating word senses in ancient Greek. In experiments we showed that GASC provides interpretable representations of the evolution of word senses, and achieves improved predictive performance compared to the state of the art. Further, we established a new framework to assess model accuracy against expert judgement. To our knowledge, no previous work has systematically compared the estimates from a statistical model to manual semantic annotations of ancient texts. 

This work can be seen as a step towards the development of richer evaluation schemes and models that can embed expert judgement. Future work could encode more structured cross-genre dependencies, or allow for change points that occur in the light of exogenous forces by historical events.

\setcounter{secnumdepth}{0}
\section{Acknowledgements}

This work was supported by The Alan Turing Institute under the EPSRC grant EP/N510129/1 and the seed funding grant SF042. 
We would also like to thank Viivi  L\"ahteenoja for her work on the annotation of the word \emph{kosmos}, Dr Lea Frermann for her advice, Prof Geoff Nicholls for his insights on the model, as well as the anonymous reviewers for their helpful comments.

\bibliography{biblio.bib}
\bibliographystyle{acl_natbib}

\end{document}